\documentclass[10pt,twocolumn,letterpaper]{article}

\usepackage{cvpr}              % To produce the CAMERA-READY version
\usepackage{microtype}

\usepackage{amsmath,amssymb,amsfonts}

\usepackage{booktabs}
\usepackage{multirow}

\usepackage{graphicx}
\usepackage{subcaption}
\usepackage{float}

\usepackage{algorithm}
\usepackage{algorithmic}

\usepackage{enumitem}

\usepackage{xspace}
\newcommand{\method}{SynCrash\xspace}

\definecolor{cvprblue}{rgb}{0.21,0.49,0.74}
\usepackage[pagebackref,breaklinks,colorlinks,allcolors=cvprblue]{hyperref}

\def\paperID{*****} % *** Enter the Paper ID here
\def\confName{CVPR}
\def\confYear{2026}

\title{SynCrash: A Multi-Stage Pipeline for Zero-Shot Accident Detection and Localization in Traffic Surveillance Video}

\author{
Arkya Jyoti Bagchi \quad
Ritul Jangir \quad
Varun Raskar\\
Indian Institute of Technology Jodhpur, India
}

\begin{document}
\maketitle
% ============================================================================
% ABSTRACT
% ============================================================================
\begin{abstract}

We present \method, a multi-stage pipeline for zero-shot accident detection, spatial localization, and collision-type classification in fixed-view CCTV surveillance video.
Our approach addresses the ACCIDENT @ CVPR 2026 Challenge, which requires predicting \emph{when} an accident occurs, \emph{where} in the frame the impact happens, and \emph{what type} of collision it is---all without access to labeled real-world training data.
The pipeline operates in three decoupled stages:
(1)~\textbf{Temporal localization} via a VideoMAEv2-giant backbone fine-tuned on CARLA-based synthetic clips with metadata-aware embeddings and dense sliding-window inference;
(2)~\textbf{Spatial localization} using YOLO for object detection combined with a physics-informed hybrid heuristic that leverages bounding-box overlap and trajectory-based reasoning to predict the impact point; and
(3)~\textbf{Collision-type classification} using a lightweight rule-based strategy derived from the number and configuration of detected vehicles.
The key insight is that temporal understanding benefits from supervised fine-tuning on synthetic data, whereas spatial understanding is better served by pretrained object detectors and physics priors that transfer naturally across domains.
\end{abstract}
% ============================================================================
% 1. INTRODUCTION
% ============================================================================
\section{Introduction}
\label{sec:intro}

Traffic accidents are rare, high-impact events whose automatic detection and understanding from surveillance video remains an open challenge \cite{liu2018future, sultani2018real, wu2026deep}.
Fixed-view CCTV cameras are ubiquitous in urban environments \cite{socha2020urban}, yet the footage they produce is characterized by low resolution, heavy compression artifacts, wide fields of view, partial occlusions, and highly variable lighting conditions \cite{sultani2018real}.
These properties make accident analysis substantially harder than in ego-centric dashcam settings \cite{chan2016anticipating, bao2020uncertainty}, where the camera is close to the action and resolution is relatively high.

The ACCIDENT @ CVPR 2026 Challenge~\cite{picek2026accident} formalizes this task as a joint prediction problem: given a CCTV clip containing a traffic accident, the system must predict
(i) the \emph{accident time} $\hat{t}$ in seconds,
(ii) the \emph{impact location} as normalized frame coordinates $(\hat{c}_x, \hat{c}_y) \in [0,1]^2$, and
(iii) the \emph{collision type} (\eg \texttt{rear-end}, \texttt{head-on}, \texttt{t-bone}).
The final score is the harmonic mean of temporal, spatial, and classification component scores, so poor performance on any single axis severely penalizes the overall result. The benchmark is training-free on real data, with only synthetic CARLA~\cite{dosovitskiy2017carla} data available, requiring zero-shot sim-to-real generalization.

Our contributions are as follows:
 A modular three-stage pipeline that decomposes accident understanding into temporal, spatial, and classification sub-problems, each addressed with the most appropriate tool.
A domain-bridging training strategy for VideoMAEv2-giant \cite{wang2023videomae} that combines Nexar weight initialization with heavy degradation augmentations to simulate CCTV artifacts.
A physics-informed spatial localization method that integrates object detection, short-term trajectory estimation, and hierarchical geometric reasoning to infer impact points without spatial supervision on real data.
An analysis of design decisions for zero-shot accident understanding, including the trade-offs between end-to-end and modular approaches.

% \input{sec/2_related}
% ============================================================================
% 3. METHOD
% ============================================================================
\section{Method}
\label{sec:method}

\subsection{Overview of SynCrash Pipeline}
\label{sec:overview}

We address accident understanding as a joint prediction problem over three components: \emph{when} the accident occurs, \emph{where} the impact happens, and \emph{what} type of collision takes place. Instead of learning these outputs jointly, we adopt a modular pipeline that decomposes the task into three stages: temporal localization, spatial localization, and collision-type classification.

Given an input CCTV video, we first perform temporal localization using a VideoMAEv2-based model~\cite{wang2023videomae} to estimate the accident timestamp $\hat{t}_{\text{acc}}$. Next, we localize the impact point by detecting objects in the corresponding frame and applying a physics-informed heuristic that reasons about spatial relationships and short-term motion. Finally, we classify the collision type using a lightweight rule-based strategy derived from the number and configuration of detected vehicles.

This design is motivated by the zero-shot sim-to-real setting of the benchmark. Temporal patterns such as motion dynamics transfer reasonably well from synthetic to real data \cite{tremblay2018training, tobin2017domain}, making them amenable to supervised learning. In contrast, precise spatial localization is highly sensitive to domain shift, and is better handled using object-centric representations and geometric reasoning. By decoupling the problem, SynCrash leverages the strengths of each component while avoiding the limitations of end-to-end multi-task models.

\subsection{Temporal Localization}
\label{sec:temporal}

We localize the accident time by framing the problem as binary classification over short video clips. We use VideoMAEv2-giant~\cite{wang2023videomae} as the temporal backbone, a ViT-based model pretrained via masked autoencoding for video understanding. Given a 16-frame clip of size $224 \times 224$, the model outputs spatiotemporal tokens that are mean-pooled to obtain a feature vector $\mathbf{v} \in \mathbb{R}^{1408}$. To account for environmental variability in CCTV footage, we incorporate categorical metadata—scene layout, weather, and time-of-day—each embedded into $\mathbb{R}^{32}$ and concatenated with $\mathbf{v}$. The combined representation is fed to a two-layer MLP with ReLU and dropout for binary prediction. The backbone is initialized with Nexar dashcam weights, providing domain-relevant priors for traffic dynamics and vehicle motion.

Training is performed on synthetic CARLA data~\cite{dosovitskiy2017carla}. From each video, overlapping 16-frame clips are extracted (stride 8) and labeled based on proximity to the ground-truth accident time $t_{\text{acc}}$:
\begin{equation}
    y =
    \begin{cases}
        1 & \text{if } t_{\text{acc}} - 1.5 \leq t_{\text{end}} \leq t_{\text{acc}} + 0.5, \\
        0 & \text{if } t_{\text{end}} < t_{\text{acc}} - 1.5,
    \end{cases}
\end{equation}
while post-accident clips are discarded. To mitigate the synthetic-to-real domain gap, we apply temporally consistent augmentations that simulate CCTV artifacts \cite{tremblay2018training}, including resolution degradation, motion blur, noise, brightness and contrast shifts, and random occlusions.

At inference, videos are processed using a dense sliding window of 16-frame clips (stride 2). Each clip is passed through the model to obtain an accident probability, and the resulting sequence is smoothed using a Gaussian filter. The predicted accident time $\hat{t}_{\text{acc}}$ is taken as the midpoint of the clip with the highest smoothed probability.

\subsection{Spatial Localization}
\label{sec:spatial}

Given the predicted accident time $\hat{t}_{\text{acc}}$, we localize the impact point $(c_x, c_y)$ in the corresponding video frame. Our approach combines object detection with physics-informed reasoning.

We use a YOLO-based \cite{redmon2016you} detector to extract vehicle bounding boxes at the frame closest to $\hat{t}_{\text{acc}}$. To incorporate motion cues, we construct short-term trajectories over a fixed window of preceding frames by matching detections using a greedy nearest-neighbor strategy similar to lightweight tracking approaches \cite{zhang2022bytetrack}, and estimate per-object velocities via linear regression over tracked positions.

The collision point is then inferred using a priority-based geometric heuristic. If any pair of objects overlaps, we use the centroid of their intersection, prioritizing pairs with larger combined area. Otherwise, we estimate the intersection of motion trajectories for approaching objects. When this is unreliable, we fall back to a size-weighted midpoint of object centers, followed by a proximity-based midpoint of the closest pair. In degenerate cases, we use the center of a single detected object, or default to the frame center $(0.5, 0.5)$ when no detections are available.

Object pairs are prioritized using an approach velocity measure:
\begin{equation}
v_{\text{approach}} = -\big[(\mathbf{c}_j - \mathbf{c}_i) \cdot (\mathbf{v}_i - \mathbf{v}_j)\big],
\end{equation}
which captures the rate at which two objects are converging. This hybrid design enables robust spatial localization under noisy detections and low-resolution CCTV conditions without requiring spatial supervision.

\subsection{Collision-Type Classification}
\label{sec:type}

The collision type is inferred using a lightweight rule-based heuristic derived from the spatial configuration and motion of detected vehicles at the predicted accident frame. We first distinguish single-vehicle events when only one object is detected, classifying them as \texttt{single}. For multi-vehicle cases, we consider relative motion cues when available: if two vehicles exhibit approximately anti-parallel motion directions, the event is classified as \texttt{head-on}; if their motion is largely aligned and one follows the other, it is classified as \texttt{rear-end}; and if their motion directions are approximately orthogonal, indicating lateral interaction, the event is classified as \texttt{t-bone}. In practice, due to noisy detections and limited temporal context, reliable motion estimation is not always available; in such cases, we fall back to a simplified object-count heuristic, mapping all multi-vehicle interactions to \texttt{t-bone}. This design maintains consistency with the spatial localization stage while providing a simple and efficient approximation of collision types under the training-free setting.

% ============================================================================
% 4. EXPERIMENTAL SETUP
% ============================================================================
\section{Experimental Setup}
\label{sec:experiments}

We follow the ACCIDENT @ CVPR 2026 Challenge protocol, where models are trained only on synthetic data and evaluated on real-world CCTV footage without access to labeled real training samples. The synthetic CARLA dataset~\cite{dosovitskiy2017carla} provides fixed-view traffic videos with annotations for accident time, impact location, and collision type, along with coarse metadata. The real test set consists of curated CCTV clips exhibiting low resolution, compression artifacts, occlusions, and diverse environmental conditions.

Performance is evaluated using the official ACCIDENT benchmark metric, which combines three components: temporal accuracy ($T$), spatial localization accuracy ($S$), and collision-type classification accuracy ($C$). Each component produces a score in $[0,1]$, and the final score is defined as their harmonic mean:
\begin{equation}
\text{Score} = \frac{3}{\frac{1}{T} + \frac{1}{S} + \frac{1}{C}}.
\end{equation}
Temporal and spatial scores are computed using Gaussian-based similarity functions over prediction errors, while classification is measured using top-1 accuracy.

\section{Results and Analysis}
\label{sec:results}

\subsection{Main Results}

Table~\ref{tab:results} presents the performance of different architectural designs on the ACCIDENT benchmark. Our proposed modular pipeline (\method{}) achieves competitive overall performance on both public and private leaderboards. Scores correspond to the official ACCIDENT metric (harmonic mean of temporal, spatial, and classification scores), reported on the public and private leaderboards. The public leaderboard is computed on a validation subset, while the private leaderboard reflects final evaluation on a held-out test set.

\begin{table}[h]
    \centering
    \caption{Comparison of different approaches on the ACCIDENT benchmark. Scores denote the official evaluation metric (harmonic mean of temporal, spatial, and classification scores), reported on the public and private leaderboards.}
    \label{tab:results}
    \small
    \begin{tabular}{@{}l|cc@{}}
        \toprule
        \textbf{Method} & \textbf{Public} & \textbf{Private} \\
        \midrule
        ViViT (joint multi-task) & 0.28 & 0.28 \\
        VideoMAEv2 + Grad-CAM + rule-based & 0.34 & 0.33 \\
        \textbf{VideoMAEv2 + YOLO + heuristic (Ours)} & \textbf{0.38} & \textbf{0.40}  \\
        VideoMAEv2 + Q-former (query-based) & 0.37 & 0.36 \\
        % Qwen (LLM-based) & 0.31 & 0.30 \\
        RAFT-based motion modeling & 0.29 & 0.28 \\
        Graph-based interaction model & 0.27 & 0.25 \\
        \bottomrule
    \end{tabular}
\end{table}

Our approach ranks 17th overall on the private leaderboard. While not the top-performing method, it provides a strong and efficient baseline under the training-free setting, demonstrating the effectiveness of combining temporal modeling with object-centric spatial reasoning.

\subsection{Effect of Design Choices}

Our final pipeline emerged through a sequence of design iterations, each addressing specific limitations observed in earlier approaches.

We initially explored a joint multi-task model based on ViViT~\cite{arnab2021vivit}, where temporal, spatial, and classification tasks were learned jointly. While it captured coarse temporal patterns, performance was poor (Table~\ref{tab:results}), with unstable spatial predictions and strong classification bias, indicating weak task disentanglement.

We therefore adopted a temporal-first pipeline, predicting accident timing before spatial reasoning. Using VideoMAEv2~\cite{wang2023videomae} significantly improved temporal localization, confirming that motion dynamics transfer well from synthetic to real data.

For spatial localization, we explored two alternatives. A Grad-CAM \cite{selvaraju2017grad}-based approach, which extracts attention maps from the temporal model, provided a lightweight solution but suffered from coarse spatial resolution and sensitivity to domain shift. In contrast, combining YOLO-based \cite{redmon2016you} object detection with physics-informed reasoning yielded more accurate and stable localization, particularly in multi-vehicle scenarios where interactions cannot be inferred from a single object. We also investigated query-based temporal reasoning using a Q-former \cite{shao2024accidentblip2}, which improved long-term context modeling but introduced higher computational cost and lacked explicit spatial grounding, limiting its effectiveness for precise localization.

Beyond these approaches, we experimented with motion-based and interaction-based models. A RAFT-based \cite{teed2020raft} pipeline leveraged dense optical flow to capture motion discontinuities associated with collisions, but was highly sensitive to noise and compression artifacts in CCTV footage. A graph-based approach modeled interactions between vehicles using trajectory features, but depended heavily on reliable tracking and was difficult to scale under noisy detections. While both approaches provided useful insights, they did not consistently outperform the simpler modular design.

Overall, these experiments highlight that decoupling temporal and spatial reasoning, and relying on  object-centric representations with physics-based heuristics, provides a more robust and practical solution under the zero-shot sim-to-real constraints of the benchmark.

\section{Discussion and Lessons Learned}
\label{sec:discussion}

\subsection{Why the Modular Design Works}

Our experiments indicate that the primary strength of \method{} lies in its modular decomposition of the problem into temporal, spatial, and classification stages. This design is particularly well-suited to the zero-shot sim-to-real setting of the benchmark, where different components of the task exhibit distinct sensitivities to domain shift.

First, temporal localization benefits from learned representations. Motion dynamics such as approaching vehicles and abrupt deceleration patterns transfer reasonably well from synthetic to real data when combined with domain-bridging augmentations. The VideoMAEv2 \cite{wang2023videomae} backbone is effective in capturing these spatiotemporal cues, enabling reliable prediction of accident timing.

In contrast,  spatial localization is highly sensitive to domain gap. Differences in camera geometry, resolution, and compression artifacts make it difficult for end-to-end models to generalize precise spatial predictions. By relying on object-centric representations (via YOLO) and physics-informed reasoning, our approach avoids learning fragile visual mappings and instead exploits geometric relationships that are largely domain-invariant.

Another key factor is the use of  specialized components. VideoMAEv2 \cite{wang2023videomae} excels at temporal understanding but provides only coarse spatial signals due to its patch-based representation. Conversely, object detectors provide accurate localization but lack temporal context. Decoupling allows each component to operate in its regime of strength without interference from competing objectives.

Finally,  physics-based reasoning provides robustness. Heuristics based on overlap, trajectory intersection, and approach velocity rely on geometric consistency rather than learned appearance features, making them more stable under noisy and low-resolution CCTV conditions.

Overall, these observations suggest that, under training-free constraints, combining learned temporal representations with object-centric detection and physics-informed reasoning offers a more reliable alternative to fully end-to-end approaches.

\subsection{What Did Not Work (and Why)}

In addition to our final pipeline, we explored several alternative approaches that, while promising in principle, did not perform reliably under the constraints of the benchmark.

 Joint multi-task learning using a shared backbone (e.g., ViViT) struggled to balance competing objectives. While temporal signals were partially captured, spatial predictions were unstable and classification exhibited strong bias toward dominant classes. This suggests that tightly coupling heterogeneous tasks under domain shift can degrade overall performance.  Grad-CAM \cite{selvaraju2017grad} based spatial localization was evaluated as a lightweight alternative that reuses the temporal model for spatial prediction. However, the coarse resolution of patch-level attention maps and their sensitivity to synthetic-to-real domain differences resulted in imprecise localization, particularly in multi-vehicle scenarios.  Query-based temporal reasoning with a Q-former \cite{shao2024accidentblip2} improved long-range temporal context modeling, but introduced higher computational overhead and lacked explicit mechanisms for spatial grounding. As a result, it was less effective for tasks requiring precise $(x,y)$ localization.  Motion-based approaches using optical flow (RAFT) \cite{teed2020raft} aimed to capture collision-specific motion discontinuities. However, these methods were highly sensitive to noise, compression artifacts, and low contrast in CCTV footage, leading to unstable motion features and inconsistent predictions.  Graph-based interaction models attempted to represent traffic scenes as dynamic graphs of interacting vehicles. While conceptually appealing, their performance depended heavily on accurate tracking and stable detections, which are difficult to obtain in low-quality surveillance videos.

Overall, these findings highlight that methods relying heavily on dense motion estimation or complex relational modeling are less robust in degraded visual conditions, and reinforce the effectiveness of simpler, modular approaches that leverage domain-invariant reasoning.

\subsection{Practical Insights for Zero-Shot Accident Detection}

Based on our experiments, we identify several practical insights for accident understanding under zero-shot sim-to-real constraints. Decoupling temporal localization, spatial prediction, and classification is critical, as these tasks exhibit different sensitivities to domain shift; separating them leads to more stable and interpretable performance than joint models. For temporal reasoning, motion cues such as vehicle approach and sudden deceleration transfer more reliably from synthetic to real data than appearance features, especially with domain-bridging augmentations. For spatial reasoning, object-centric representations are more effective, as explicit detections provide a reliable basis for modeling interactions compared to implicit attention maps. Physics-based cues, including overlap, trajectory intersection, and approach velocity, generalize well across domains and remain robust under noise. In contrast, methods relying heavily on dense motion estimation or complex relational modeling are sensitive to noise and require high-quality inputs, limiting their effectiveness in low-resolution CCTV settings. Overall, combining learned temporal representations with simple, interpretable, physics-informed spatial reasoning provides a practical and robust solution under limited supervision.

\section{Limitations}
\label{sec:limitations}

Our approach has several limitations. Collision-type classification relies on simple heuristics and struggles with fine-grained multi-vehicle interactions, especially under noisy detections. Spatial localization depends on detection quality, making it vulnerable in crowded or low-resolution scenes. Temporal localization is limited by fixed-length clips, reducing sensitivity to very short or gradual events. The pipeline is also sequential, with no feedback between stages, limiting joint optimization. While robust to sim-to-real conditions, the method does not explicitly perform domain adaptation and may degrade under extreme lighting, weather, or viewpoints. Improving interaction modeling, detection robustness, and tighter stage integration are important directions for future work.
{
    \small
    \bibliographystyle{ieeenat_fullname}
    \bibliography{main}
}

\end{document}